\documentclass{article}

\usepackage{arxiv}

\usepackage{multicol} 

\usepackage{amsmath}
\usepackage{amssymb}

\usepackage[utf8]{inputenc} 
\usepackage[T1]{fontenc}    
\usepackage{hyperref}       
\usepackage{url}            
\usepackage{booktabs}       
\usepackage{amsfonts}       
\usepackage{nicefrac}       
\usepackage{microtype}      
\usepackage{cleveref}       
\usepackage{lipsum}         
\usepackage{graphicx}
\usepackage[square,numbers,super,sort&compress]{natbib}
\usepackage{doi}

\usepackage{tikz}                    
\usetikzlibrary{shapes, arrows, positioning, fit, backgrounds, calc, matrix}
\usepackage{caption}                 
\usepackage{float}                   
\usepackage{geometry}                
\usepackage{adjustbox} 
\usetikzlibrary{arrows.meta}  

\usepackage{rotating}  
\usepackage{pdflscape} 

\usepackage{booktabs, threeparttable} 
\usepackage{longtable}
\usepackage{tabularx}
\usepackage[justification=centerlast]{caption} 
\usepackage{color}

\crefname{section}{Sec.}{Secs.}
\Crefname{section}{Section}{Sections}
\Crefname{table}{Table}{Tables}
\crefname{table}{Tab.}{Tabs.}

\usepackage{pgfplots}
\pgfplotsset{compat=1.18}

\usepackage{listings} 

\definecolor{codegreen}{rgb}{0,0.6,0}
\definecolor{codegray}{rgb}{0.5,0.5,0.5}
\definecolor{codepurple}{rgb}{0.58,0,0.82}
\definecolor{backcolour}{rgb}{0.95,0.95,0.92}

\lstdefinestyle{promptstyle}{
	backgroundcolor=\color{backcolour},   
	commentstyle=\color{codegreen},
	keywordstyle=\color{magenta},
	numberstyle=\tiny\color{codegray},
	stringstyle=\color{codepurple},
	basicstyle=\ttfamily\footnotesize,
	breakatwhitespace=false,         
	breaklines=true,                 
	captionpos=b,                    
	keepspaces=true,                 
	numbers=left,                    
	numbersep=5pt,                  
	showspaces=false,                
	showstringspaces=false,
	showtabs=false,                  
	tabsize=2,
	frame=single,
	rulecolor=\color{black}
} 

\lstdefinestyle{pythonstyle}{
	backgroundcolor=\color{backcolour},   
	commentstyle=\color{codegreen},
	keywordstyle=\color{magenta},
	numberstyle=\tiny\color{codegray},
	stringstyle=\color{codepurple},
	basicstyle=\ttfamily\footnotesize,
	breakatwhitespace=false,         
	breaklines=true,                 
	captionpos=b,                    
	keepspaces=true,                 
	numbers=left,                    
	numbersep=5pt,                  
	showspaces=false,                
	showstringspaces=false,
	showtabs=false,                  
	tabsize=2,
	frame=single,
	rulecolor=\color{black},
	language=Python
}

\newcommand{\minitrend}[3]{%
	\begin{tikzpicture}[baseline={(0,0)}, x=0.4cm, y=0.15cm]
		\pgfmathsetmacro{\maxval}{max(#1,#2,#3)+0.0001}
		\pgfmathsetmacro{\minval}{min(#1,#2,#3)}
		\pgfmathsetmacro{\range}{\maxval-\minval}
		\pgfmathsetmacro{\yone}{(#1-\minval)/\range}
		\pgfmathsetmacro{\ytwo}{(#2-\minval)/\range}
		\pgfmathsetmacro{\ythree}{(#3-\minval)/\range}
		
		\draw[teal, thick] (0,\yone) -- (1,\ytwo) -- (2,\ythree);
		\fill[teal] (0,\yone) circle (1.2pt);
		\fill[teal] (1,\ytwo) circle (1.2pt);
		\fill[teal] (2,\ythree) circle (1.2pt);
	\end{tikzpicture}%
}

\newcommand{\minitrendLogd}[3]{%
	\begin{tikzpicture}[baseline={(0,0)}, x=0.4cm, y=0.15cm]
		\ifdim\range pt=0pt
			\draw[blue, thick] (0,1) -- (2,1);
			\fill[blue] (0,1) circle (1.2pt);
			\fill[blue] (1,1) circle (1.2pt);
			\fill[blue] (2,1) circle (1.2pt);
		\else
			\pgfmathsetmacro{\yone}{ln(#1+1)/ln(10)}
			\pgfmathsetmacro{\ytwo}{ln(#2+1)/ln(10)}
			\pgfmathsetmacro{\ythree}{ln(#3+1)/ln(10)}
			
			\pgfmathsetmacro{\maxlog}{max(\yone,\ytwo,\ythree)}
			\pgfmathsetmacro{\scaledone}{\yone/\maxlog*2}
			\pgfmathsetmacro{\scaledtwo}{\ytwo/\maxlog*2}
			\pgfmathsetmacro{\scaledthree}{\ythree/\maxlog*2}
			
			\draw[gray!30, very thin] (0,0) rectangle (2.5,2.5);
			
			\draw[blue, thick] (0,\scaledone) -- (1,\scaledtwo) -- (2,\scaledthree);
			\fill[blue] (0,\scaledone) circle (1.2pt);
			\fill[blue] (1,\scaledtwo) circle (1.2pt);
			\fill[blue] (2,\scaledthree) circle (1.2pt);
		\fi
	\end{tikzpicture}%
}

\newcommand{\dpaperbib}{tcm-radiotherapy-25yrs}

\newcommand*{\zsubsection}{\subsection} 
\newcommand*{\zsubsubsection}{\subsubsection}

\newcommand{\PMIDzFetchDate}{07-March-2025}

\newcommand{\NPMIDZ}{69,745}
\newcommand{\NPMIDZLLM}{4,078}
\newcommand{\NPMIDZLLMPERCENT}{5.8\%}

\newcommand{\NPMIDZLastYEAR}{9,407}
\newcommand{\NPMIDZLastYEARPERCENT}{13.5\%}

\newcommand{\NPMIDZLastTenYEARSPERCENT}{64.9\%}

\newcommand{\PercAnnualGrowthPublications}{$6.3\%$}

\newcommand{\TOPk}{30}

\newcommand{\NTOPKJournalsPERCENT}{11.3\%}

\newcommand{\NPMIDZLLMVALIDATED}{2113} 
\newcommand{\NPMIDZLLMVALIDATEDPERCENTALL}{3.0\%}
  
\newcommand{\NOBSLLMVALIDATED}{5,363}

\newcommand{\lblBasicResearch}{basic research}

\newcommand{\EpochNascent}{Low-alignment}
\newcommand{\EpochFirstWave}{First wave}
\newcommand{\EpochSecondWave}{Second wave}

\newcommand{\TScopeComplications}{Supportive care and complication managment} 
\newcommand{\TScopeCTherapeutics}{Therapeutics and mechanisms} 
\newcommand{\TScopeCancerTypes}{Cancer type} 
\newcommand{\TScopeOutcomes}{Clinical endpoints and planning} 
\newcommand{\TScopeOthers}{Others}

\newcommand{\ThemesCoreCAP}{Core themes}
\newcommand{\ThemesBrokersCAP}{Bridging themes}
\newcommand{\ThemesNicheCAP}{Peripheral themes}
\newcommand{\ThemesCore}{core themes}
\newcommand{\ThemesBrokers}{bridging themes}
\newcommand{\ThemesNiche}{peripheral themes}

\newcommand{\QWEN}{QWen2.5}

\newcommand{\LLMClassifierFULL}{QWen2.5 32B}

\newcommand{\BERTFull}{All-MiniLM-L6-v2}

\newcommand{\LLMClassifier}{QWen}

\newcommand{\lblSuccessful}{positive} 
\newcommand{\lblUnSuccessful}{negative}  
\newcommand{\lblMixed}{mixed} 
\newcommand{\lblUnclear}{unclear} 

\newcommand{\nPermutationsThemesActual}{360}
\newcommand{\JaccardThreshold}{$98.5\%$}

\newcommand{\colorCoreThemez}{green}
\newcommand{\colorNicheThemez}{yellow}

\newcommand{\TheTitle}{Mapping the maturation of TCM as an adjuvant to radiotherapy}
\newcommand{\TheTitleShort}{\TheTitle}

\usepackage{caption}  
\usepackage{subcaption} 

\title{\TheTitle}

\date{}

\newif\ifuniqueAffiliation
 
\usepackage{authblk}

\newbox{\orcid}\sbox{\orcid}{\includegraphics[scale=0.06]{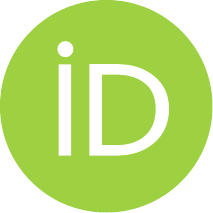}} 
\author[1]{%
	\href{https://orcid.org/0009-0000-2080-4979}{\usebox{\orcid}\hspace{1mm}P.~Bilha Githinji}%
}
\author[1]{%
	\href{https://orcid.org/0009-0001-3472-4678}{\usebox{\orcid}\hspace{1mm}Aikaterini Melliou}%
}  

\author[1]{%
	\href{}{\usebox{\orcid}\hspace{1mm}Xi Yuan}%
}

\author[1]{%
	\href{}{\usebox{\orcid}\hspace{1mm}Dayan Zhang}%
}

\author[1]{%
	\href{}{\usebox{\orcid}\hspace{1mm}Lian Zhang}%
}

\author[1]{%
	\href{}{\usebox{\orcid}\hspace{1mm}Zhenglin Chen}%
}

\author[1]{%
	\href{}{\usebox{\orcid}\hspace{1mm}Jiansong Ji}%
}

\author[1]{%
	\href{}{\usebox{\orcid}\hspace{1mm}Chengying Lv}%
}

\author[1]{%
	\href{}{\usebox{\orcid}\hspace{1mm}Jinhao Xu}%
}

\author[1]{%
	\href{https://orcid.org/0000-0001-7336-7848}{\usebox{\orcid}\hspace{1mm}Peiwu Qin}%
}

\author[2]{%
	\href{}{\usebox{\orcid}\hspace{1mm}Dongmei Yu}%
}

\affil[1]{Guangdong Provincial Laboratory of Traditional Chinese Medicine, Hengqin, Guangdong, China}
\affil[2]{Affiliated Fifth Hospital, Wenzhou Medical University, Wenzhou, Zhejiang, China}

\renewcommand{\headeright}{} 
\renewcommand{\undertitle}{} 
\renewcommand{\shorttitle}{\TheTitleShort}

\hypersetup{
	pdftitle={\TheTitle},
	pdfsubject={q-bio.NC, q-bio.QM},
	pdfauthor={P. Bilha.~Githinji, Et al.},
	pdfkeywords={TCM, Radiotherapy, Integrative Oncology, NLP, Global Landscape},
}

\begin{document}
\maketitle

\begin{abstract}	

The integration of complementary medicine into oncology represents a paradigm shift that has seen to increasing adoption of Traditional Chinese Medicine (TCM) as an adjuvant to radiotherapy. 
About twenty-five years since the formal institutionalization of integrated oncology, it is opportune to synthesize the trajectory of evidence for TCM as an adjuvant to radiotherapy.
Here we conduct a large-scale analysis of~\NPMIDZ~publications (2000 - 2025), emerging a cyclical evolution defined by coordinated expansion and contraction in publication output, international collaboration, and funding commitments that mirrors a \emph{define-ideate-test} pattern. 
Using a theme modeling workflow designed to determine a stable thematic structure of the field, we identify five dominant thematic axes - cancer types, supportive care, clinical endpoints, mechanisms, and methodology -  that signal a focus on patient well-being, scientific rigor and mechanistic exploration. Cross-theme integration of TCM is patient-centered and systems-oriented. Together with the emergent cycles of evolution, the thematic structure demonstrates progressive specialization and potential defragmentation of the field or saturation of existing research agenda. 
The analysis points to a field that has matured its current research agenda and is likely at the cusp of something new. 
Additionally, the field exhibits positive reporting of findings that is homogeneous across publication types, thematic areas, and the cycles of evolution suggesting a system-wide positive reporting bias agnostic to structural drivers.

	\keywords{TCM \and Radiotherapy \and Integrative Oncology \and Global Landscape}
\end{abstract}

\section{Introduction}
\label{sec:dpintro}

Traditional Chinese Medicine (TCM) is increasingly an important adjunctive therapy alongside radiotherapy within integrative oncology practice and research, and is frequently utilized by patients for relief from cancer- and treatment-related symptom burden~\cite{lapen_use_2021,mao_integrative_2022,bai_bibliometric_2023}. 
Over the past quarter-century, the literature has expanded to encompass a growing body of evidence addressing treatment tolerance and adherence, symptom burden, and patient-reported outcomes and experiences alongside preclinical and mechanistic investigations~\cite{lapen_use_2021,mao_integrative_2022,bai_bibliometric_2023,gowin_integrative_2024}. 
This maturation creates a timely opportunity for consolidation and macro-level synthesis of the field's intellectual foundation and temporal dynamics, informing future research directions and evidence synthesis priorities.

To conduct such a synthesis, advances in computational bibliometric analysis and natural language processing (NLP) text analysis enable fine-grained characterization of large-scale research landscapes and facilitate automated semantic interrogation of scientific texts across large corpora. 
One such opportunity for semantic-based analysis concerns how results are framed and communicated (rhetorical framing of results), offering a lens on field-level optimism and potential systemic incentives and structural drivers of reporting behavior~\cite{vinkers_use_2015,chiu_spin_2017,duyx_strong_2019}.  
Conversely, inherent stochasticity in these computational tools and particularly in topic modeling or clustering-based theme extraction, can lead to inconsistency in inferred thematic structures, hindering reproducibility and complicating interpretability~\cite{hosseiny_marani_review_2024}.

In this study, we analyze a large corpus of~\NPMIDZ~publications examining TCM as an adjunctive therapy to radiotherapy over the past twenty-five years. 
We construct a stable latent thematic structure of the field through a 98.5\% Jaccard similarity-based convergence across multiple topic-modeling instantiations that employ different text embedding models and clustering hyperparameters for term grouping. 
Furthermore, utilizing a large language model (LLM)-based pipeline for extraction and classification, we examine the rhetorical framing of results reported in the abstracts.


Specific contributions of this report include 
\begin{itemize}
	
	\item{We map the field across a wider range of publication types and cancer types with a corpus of~\NPMIDZ~publications. }
	
	\item{We empirically emerge a cyclic  pattern of evolution in the field that is based on congruence of growth patterns across multiple activity indicators concerning publication output, collaboration breadth, and funding acknowledgments. This cyclic pattern  is further evidenced in the the thematic structure and, together, these two empirical results portray a field that appears to follow a \emph{define-ideate-test} pattern so far. From the observed pattern, the years 2025 - 2027 could be at the cusp of the next interval of reflection that precedes a multi-year period of expansion.}
	
	\item{We find that while there is rhetorical positive reporting of findings within abstract texts in the field, this reporting bias is homogeneous across publication types, thematic areas, and the identified cycles of evolution. }

	\item{Methodologically, we establish a stable, convergent thematic structure of the field. By employing modern NLP embedding for topic modeling, operating on the entire abstract document, and converging topic modeling outputs from multiple algorithmic ensembles of embedding models and clustering hyperparameters, we identify a consensus thematic structure that has richer semantic breadth than is usually achievable with keywords analysis, and that is determined with a Jaccard similarity threshold of 98.5\%. 
	}
	 
\end{itemize}

\section{Background}
\label{sec:dpbackground}

Radiotherapy is a central component of modern oncology, fulfilling definitive, adjuvant and palliative roles across a wide range of cancers, and featuring in the treatment course of more than half of all cancer patients~\cite{borm_complementary_2020}.
However, radiotherapy is also associated with treatment-related toxicity. Patients commonly experience cancer-related and radiation-induced effects like fatigue, nausea and vomiting, dermatitis, pain, sleep disturbance, anxiety, and depressive symptoms, with some adverse effects persisting months or years after treatment~\cite{borm_complementary_2020,mao_integrative_2022,lapen_use_2021,gowin_integrative_2024}.
These complications can markedly reduce quality of life and lead to treatment interruptions, compromising clinical outcomes.

The positioning of TCM as a complementary or adjunctive therapy rather than an alternative modality in oncology was driven by sustained patient demand and the institutionalization of integrative oncology as a clinical and research discipline~\cite{mao_integrative_2022,gowin_integrative_2024,lyman_integrative_2018}. 
Integrative oncology provided a conceptual, clinical and organizational framework through which traditional practices could be systematically defined, evaluated using biomedical research methods, and clinically integrated alongside conventional cancer treatments~\cite{mao_integrative_2022,gowin_integrative_2024,cramer_integrative_2013,witt_comprehensive_2017}.  
Its  central aim is to combine conventional and complementary approaches in a manner that prioritizes safety, clinical effectiveness, and patient-centered care~\cite{gowin_integrative_2024,cramer_integrative_2013,witt_comprehensive_2017}. 
In TCM, syndrome and the modern definition of disease are conceptualized as patterns (zheng) of systemic disharmony, and treatment aims at regulating qi, blood, yin-yang balance and organ systems. 
TCM represents a major component of complementary medicine within integrative oncology, and includes modalities such as Chinese herbal medicine (formula-based prescriptions), TCM dietary therapy (shiliao), acupuncture and moxibustion, and qigong (mind-body exercise)~\cite{world_health_organization_who_2013,world_health_organization_who_2019}.

Recent bibliometric studies mapping TCM-related research in oncology have restricted their scope to specific cancer types, individual TCM modalities or narrowly defined clinical or biological contexts. 
Cancer-specific bibliometric analyses commonly identify mechanism-focused topics as dominant themes. In lung cancer, anticancer mechanisms and computational and systems-level approaches such as network pharmacology and molecular docking feature, while liver cancer studies highlight apoptosis and signaling pathways such as NF-$\kappa$B~\cite{qin_visual_2025,shi_effectiveness_2025}.
Similarly, context-constrained analyses such as those examining immune modulation or gut microbiota regulation, discover themes on pharmacodynamics, immune checkpoints, signaling pathway regulation, apoptosis, short-chain fatty acids, and computational multi-omics techniques~\cite{lei_bibliometric_2025,lian_bibliometric_2025,lei_bibliometric_2025-1}.
Conversely, bibliometric analyses on clinical application of individual TCM modalities in supportive or adjunctive care report patient and outcome-centered themes. For example, acupuncture-focused analyses identify themes regarding management of treatment-related or cancer-related symptoms (e.g. pain, fatigue, sleep disturbance) and improvement in health-related quality of life of the patient~\cite{liu_hotspots_2024,shang_current_2024}.
The analytical framing of these studies seems to impact the discovered themes.

Broadly scoped bibliometric studies examining TCM-related research in oncology likewise identify cellular and molecular-level investigations, preclinical experimental methodologies, anticancer mechanisms, and pharmacokinetic (PK) and pharmacodynamic (PD) analyses as dominant themes~\cite{bai_bibliometric_2023,zhang_evolution_2023,akalin_traditional_2025}.
Although these studies avoid narrow topical constraints in their search strategies, their analytical designs introduce implicit selection and interpretive limitations. 
For example, one study restricts analysis to publications affiliated with Chinese academic institutions~\cite{zhang_evolution_2023}, while others limit inclusion to article and review publication types, thereby disproportionately weighting basic and preclinical research outputs~\cite{bai_bibliometric_2023}. 
These design choices, while methodologically defensible within their stated aims, systematically shape the thematic signal that emerges from the data.

A further defining feature of the bibliometric studies in this domain is the central role of visualization of co-word or co-occurrence networks~\cite{lei_bibliometric_2025,lian_bibliometric_2025,bai_bibliometric_2023,zhang_evolution_2023,hosseiny_marani_review_2024}.
While these visual tools are valuable for exploratory insight, their construction and iterative refinement inevitably incorporate analyst-dependent decision and potential stochastic variability and sampling artifacts inherent to clustering-based topic modeling algorithms.  
Minor variations in hyperparameter settings can lead to markedly different thematic maps and visualizations, materially altering inferred thematic structures and complicating reproducibility, robustness assessment, and interpretability~\cite{hosseiny_marani_review_2024}.

Moreover, the abstract text in bibliometric metadata is an information-dense representation of research outputs that can be leveraged, enabling examination of the rhetorical framing of scientific findings for quantifiable indicators of research and publication practices.
Systematic analysis of selective outcome reporting and rhetorical framing of results, for instance, allows empirical assessment of how publication incentives and researcher-level cognitive and motivational biases manifest in published literature, and how these patterns propagate into evidence synthesis and downstream application~\cite{kavvoura_selection_2007,vinkers_use_2015,duyx_strong_2019,chiu_spin_2017}. For example, prior work has shown that selective positive reporting can lead to systematic inflation of pooled effect size estimates in meta-analyses when unfavorable results, reported in the full text, are omitted from abstracts~\cite{duyx_strong_2019}. 
While abstracts have been shown to preferentially present positive or positively-framed results~\cite{kavvoura_selection_2007,vinkers_use_2015}, they play a crucial role in literature discoverability and preliminary study appraisal. Advances in large-scale text mining and natural language processing methods, including transformer-based language models, now enable automated semantic and discourse-level analysis of scientific texts~\cite{TODO:}. Therefore, incorporating semantic analysis of abstract texts in metascience assessments can offer a means to detect and monitor systemic reporting patterns and biases.
Moreover, joint analysis of thematic structure and rhetorical framing of findings
could emerge how thematic issues align with, or diverge from, the rhetoric used to communicate them.

This study presents a unified view of the field of TCM as an adjunctive therapy to radiotherapy over the last 25 years, without restrictions on tumor sites, cancer types, clinical contexts, TCM interventions or publication types. 
Using a comprehensive PubMed search strategy supplemented by reference tracking in PubMed Central, we assemble a final corpus of~\NPMIDZ~publications. 
To characterize the latent conceptual structure of the literature, we implement a theme-extraction pipeline that derives a robust thematic solution from consensus across multiple topic-modeling runs (Jaccard score threshold of 98.5\%), varying both text embedding models and clustering hyperparameters. 
Additionally, we employ a large language model to extract and classify result-bearing statements in abstracts, allowing us to examine rhetorical framing patterns in relation to temporal trends, study types, and thematic composition. 
In the next sections we present our results and later detail the methodological aspects of the study.

\section{Results}
\label{sec:dresults}


This study integrates analytical pipelines that merge conventional bibliometrics with detailed network analysis and recent advances in natural language processing. 
We first retrieve publication metadata from PubMed via NCBI EDirect API, yielding a primary dataset of~\NPMIDZ~records after cleaning and applying eligibility criteria.  
Subsequent analysis entails quantifying annual production, collaboration indicators and funding-related metadata, and examining longitudinal growth patterns. 
Additionally, we determine a consensus-derived set of thematic areas within the literature using a multi-stage pipeline that applies embedding-based semantic topic modeling, and aggregates multiple stochastic topic-modeling outputs using set-based similarity measures. 
We further examine thematic co-occurrence structure and its temporal evolution 
using network analysis metrics. 
Finally, we employ an LLM-based information extraction and classification pipeline to characterize the rhetorical framing of reported findings across structural dimensions.

Here, we present the results from this quantitative assessment of research examining TCM as an adjunct to radiotherapy. 
We first describe the composition of both the full bibliometric corpus and the analytical sampled subset used for abstract-level classification of the rhetorical framing of reported findings. Thereafter, we present notable trends in publication output and indicators of research engagement (authorship, collaboration, funding), and conclude with exploration of the embedding-derived thematic structure of the literature and the rhetorical success status of abstract-level reported findings.


\zsubsection{Sample Overview}
Our search process realized~\NPMIDZ~records published between 01-January-2000 and ~\PMIDzFetchDate, capturing a 25-year longitudinal publication window. 
This dataset provides sufficient scale for mapping the publication dynamics, collaboration networks and thematic composition across the field. 
Research on TCM as an adjunctive modal to radiotherapy exhibits marked growth in publication volume, with the last decade accounting for~\NPMIDZLastTenYEARSPERCENT~of the identified research, and the last year (2024) alone contributing~\NPMIDZLastYEARPERCENT~of total output (\NPMIDZLastYEAR~articles).

For assessment of the rhetorical framing of results, we draw a random sample to classify the rhetorical success status (reporting valence) of abstract-level statements of findings. A total of~\NPMIDZLLM~ articles (\NPMIDZLLMPERCENT~of the corpus) is successfully processed by the LLM and available for further analysis. As detailed in~\Cref{tbl:sample}, this subset 
is comparable to the full corpus across key bibliometric indicators on
reported funding acknowledgments, number of collaborating parties, and coverage of the analysis period. 
It is suitable for assessing the distribution of rhetorical framing of abstract-level result statements.

\begin{table}[htbp!]
\centering
\caption{Sample size and bibliometric indicators}
\label{tbl:sample}
\begin{tabular}{p{4.1cm}p{2.1cm}p{2.1cm}p{2.1cm}p{2.1cm}p{2.1cm}}
\toprule
  & Entire corpus & Last Decade & Last Year & LLM Classified & LLM Verified \\
\midrule
Publications & 69,745 & 45,284 & 9,407 & 4,078 & 2,113 \\
\% of corpus & 100.0\% & 64.9\% & 13.5\% & 5.8\% & 3.0\% \\
Period & 2000 - 2025 & 2015 - 2025 & 2024 - 2025 & 2000 - 2025 & 2000 - 2025 \\
$n$ years & 25 & 10 & 1 & 25 & 25 \\
Mean authors/paper & 6.83 & 7.43 & 7.79 & 6.86 & 6.77 \\
Mean affiliations/paper & 3.27 & 4.38 & 4.75 & 3.37 & 3.30 \\
Mean countries/paper & 1.27 & 1.42 & 1.44 & 1.28 & 1.26 \\
Mean first author active years & 1.87 & 1.79 & 1.70 & 1.89 & 1.93 \\
$n$ funding acknowledgments & 37,783 & 26,399 & 6,706 & 2,244 & 1,157 \\
Mean funding acknowledgments/paper & 0.54 & 0.58 & 0.71 & 0.55 & 0.55 \\
\bottomrule
\end{tabular}
\end{table}

\begin{table}[htbp!]
\centering
\caption{Properties of the identified epochs}
\label{tbl:sample-epochs}
\begin{tabular}{p{4.1cm}p{2.1cm}p{2.1cm}p{2.1cm}p{2.1cm}}
\toprule
  & Low-Alignment & First Wave & Second Wave & Trend \\
\midrule
Publications & 10,702 & 23,711 & 35,332 & \minitrend{4.029505525426577}{4.374968186299234}{4.5481805134080275} \\
\% of corpus & 15.3\% & 34.0\% & 50.7\% &  \\
Period & 2000 - 2008 & 2009 - 2017 & 2018 - 2025 &  \\
$n$ years & 9 & 9 & 7 &  \\
Mean authors/paper & 5.27 & 6.43 & 7.57 & \minitrend{5.27}{6.43}{7.57} \\
Mean affiliations/paper & 0.99 & 2.37 & 4.57 & \minitrend{0.99}{2.37}{4.57} \\
Mean countries/paper & 0.98 & 1.14 & 1.44 & \minitrend{0.98}{1.14}{1.44} \\
Mean first author active years & 2.00 & 2.01 & 1.74 & \minitrend{2.0}{2.01}{1.74} \\
$n$ funding acknowledgments & 3,541 & 12,458 & 21,784 & \minitrend{3.549248556854056}{4.095483185829541}{4.338157564271773} \\
Mean funding acknowledgments/paper & 0.33 & 0.53 & 0.62 & \minitrend{0.33}{0.53}{0.62} \\
\bottomrule
\end{tabular}
\end{table}

\begin{figure}[htbp!]  
    \centering  
    \includegraphics[width=0.95\textwidth]{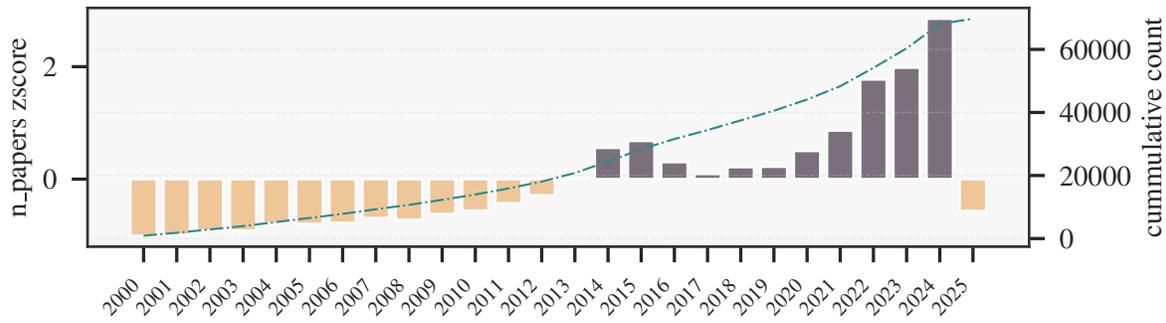}
    \caption{Annual publication output}
    \label{fig:output-pub}
    \end{figure}

\zsubsection{Research activity and growth dynamics} 
Our analysis examined publication volume, collaboration patterns, and grant acknowledgments per annum and per publication where appropriate. Collaboration metrics included the number of authors, contributing institutions, and international co-authorship. 
Longitudinal trends in publication output are illustrated in~\Cref{fig:output-pub}. The cumulative publication trajectory, plotted on the secondary y-axis, shows a continual rise in research output. Moreover, the z-score-standardized annual publication output on the primary y-axis, presents a normalized view of the year-to-year variation with years exceeding the long-term mean over the study period depicted by dark shading, and suggesting a potential inflection point in 2013. From 2014 onward, the field has experienced annual output that consistently surpasses the historical average.

Additionally,~\Cref{tbl:sample} tabulates per paper averages for reported grant support and collaborating entities (authors, institutions, countries). The corpus documents $37,783$ grants, corresponding to a mean of $0.54$ grant acknowledgments per publication; approximately one acknowledged grant per two publications. 
Each publication involves an average of $6.83$ authors drawn from $3.27$ institutions across $1.27$ countries.

\begin{figure}[htbp!]  
    \centering  
    \includegraphics[width=0.95\textwidth]{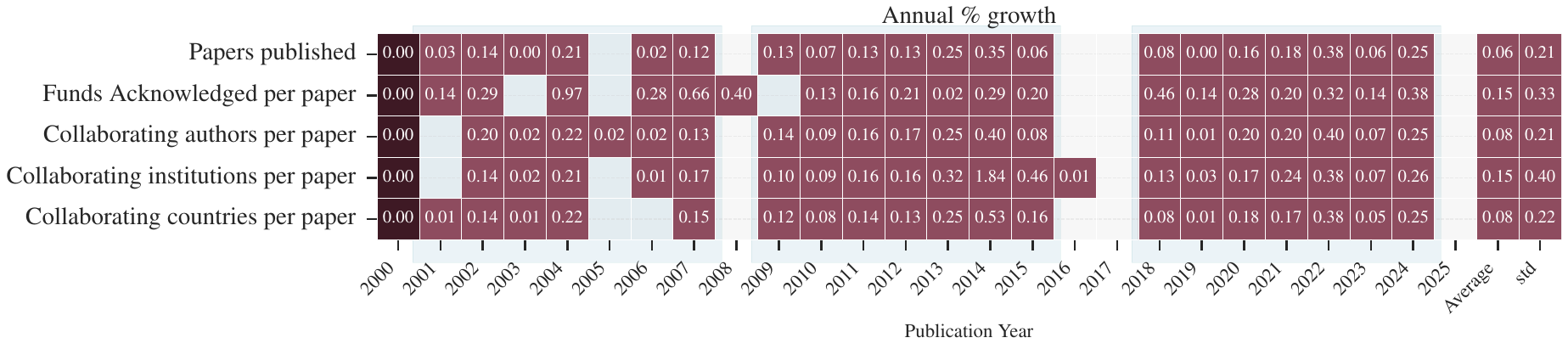}
    \caption{Growth rate of various indicators}
    \label{fig:output-growth}
    \end{figure}

\paragraph{Year-on-year growth dynamics}
We further assessed the temporal dynamics of research output and engagement by examining the year-on-year growth across multiple bibliometric indicators. 
~\Cref{fig:output-growth}, is a heatmap plotting the indicators on the y-axis and the publication years on the x-axis, with each cell denoting the proportional change from the preceding year
($\frac{ x_{\mathrm{current_year}} - x_{\mathrm{previous_year}} }{ x_{\mathrm{previous_year}} }$).
Color gradients distinguish positive growth from zero or negative growth, with the year 2000 serving as the baseline. 
The visualization reveals recurrent multi-year periods of sustained positive growth such that there are blocks of consecutive years with positive growth across most or all indicators interspersed with shorter intervals of non-positive growth. Over the past decade, for example, 2016 - 2017 are characterized by predominantly non-positive growth, followed by a sustained phase of expansion from 2018 to 2024. 
These alternating intervals are consistent with a cyclical trajectory in the development of the field, that is marked by coordinated shifts in publication output, grant acknowledgments, and collaboration indicators. This suggests interdependence among these bibliometric dimensions.

Building on the observed cycles of coordinated growth, we delineate three empirically derived developmental epochs that capture the temporal structure of the evolution of the research field. 

\begin{itemize}
	
	\item{\textbf{Low alignment (2000 - 2008).}
		This initial phase is characterized by limited synchrony across the indicators, particularly during 2001 - 2004 and 2006 - 2007. 
		Given the low absolute publication volumes in these early years, we consolidate these intervals, including the transitional period of non-positive growth in 2005, into a single nine-year epoch from 2000 to 2008. 
		This period accounts for $15.3\%$ of total research output.
	}
	
	\item{\textbf{First wave (2009 - 2017)}		
		This is the first phase of consistent coordinated growth. 
		This epoch reflects the first sustained block of coordinated expansion across publication output, reported grant activity and collaboration. Spanning seven years of predominantly positive year-on-year growth and concluding with the contraction in years 2016 and 2017, this period is nine years long and contributes $34.0\%$ of total output. 
	}
	
	\item{\textbf{Second wave (2018 - 2025)}
		This is the second phase of consistent coordinated growth. 
		The most recent epoch captures $50.7\%$ of all publications and represents the most active phase to date, with the highest observed publication activity. 
		Now seven years in, this wave encompasses the period of analysis and may be approaching a potential inflection point conditional on persistence of previously observed growth cycles.
	}
\end{itemize}

Summaries of the research output and engagement indicators for each of these epochs are presented in~\Cref{tbl:sample}. There is a monotonic increase in publication output, reported grant activity and collaborative breadth across successive developmental epochs. 
Research output has increased from $15.3\%$ to $50.7\%$ of the analysis corpus from the low alignment period to the most recent second wave epoch. 
The mean number of grant acknowledgments per publication has increased from $0.33$ during the low alignment epoch to $0.62$ during the second wave, while mean number of collaborating authors has risen from a mean of $5.27$ to $7.57$ authors per publication.

\begin{figure}[htbp!]
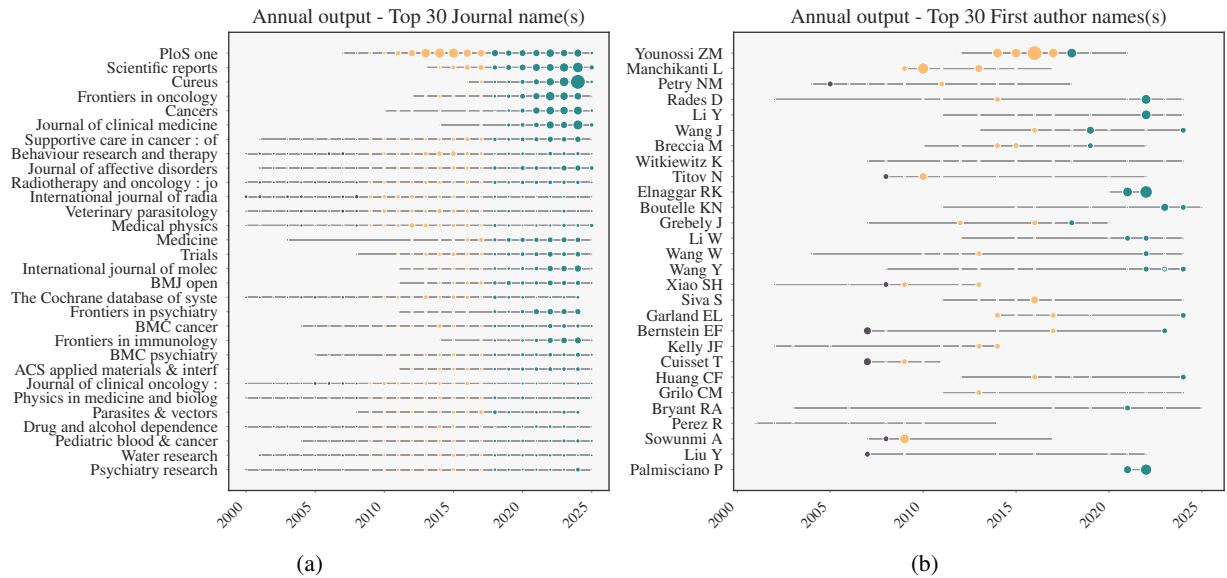
  
    \centering  
    \begin{subfigure}[b]{0.49\linewidth}
        \centering \raisebox{-\height}{\includegraphics[width=\linewidth]{output\_indicators\_\_journal\_name-n\_paperszscore}}
\caption{}
\label{fig:player-journals}
        \end{subfigure}
\begin{subfigure}[b]{0.49\linewidth}
        \centering \raisebox{-\height}{\includegraphics[width=\linewidth]{output\_indicators\_\_first\_author\_names-n\_paperszscore}}
\caption{}
\label{fig:player-authors}
        \end{subfigure}
\hfill
    \caption{Annual publication output of top journals and first authors}
    \label{fig:output-players}
    \end{figure}

\paragraph{Collaboration structure and publication venues}
Following the observed rise in collaborative activity, we detail the publication activity of leading contributors and principal publication venues. 
~\Cref{fig:player-journals} highlights the journals with the highest publication output, ranking them (top to bottom) on the y-axis according to total output, and visualizing their temporal activity using bubble size. Together, the top~\TOPk~journals account for~\NTOPKJournalsPERCENT~of identified publications.

A dual structure emerges. The top six journals are relatively recent entrants to research on TCM as an adjuvantive to radiotherapy, with activity concentrated in the latter years of the dataset. These journals, - PLoS ONE, Scientific Reports, Cureus, Fronteirs in Oncology, Cancers, and the Journal of Clinical Medicine - are broad-scope, multidisciplinary biomedical journals, suggesting that the TCM-adjuvant research has expanded beyond niche complementary medicine outlets and is increasingly visible within mainstream biomedical publishing platforms. 
Conversely, the next seven journals exhibit sustained output over a longer period of time and concentrate on specific clinical or methodological domains, including clinical oncology, radiotherapy physics and journals focused on radiotherapy effects, oncologu therapeutics or treatment response. 
This second band reflects sustained engagement from specialist research communities and signals methodological emphasis within the field.

Similarly, annual publication trajectories for the most prolific first authors by publication count are shown in ~\Cref{fig:player-authors}.These authors have contributed to the field over extended periods, mean active publication duration 
of $13.87$ years. 
Similar visualizations for leading institutions and countries are provided in~\Cref{sec:appendix-results}.
At the country level, the~\TOPk~contributors demonstrate persistent activity across the entire 25-year window, indicating broad and temporally stable international research participation, 
with TCM as an adjuvant approach in oncology. 
The United States (USA) leads publication output until 2021, 
after which China incrementally surpasses USA publication output.

\begin{table}[htbp!]
\centering
\caption{Publication output of top contributors and publication types}
\label{tbl:summary-stats-grouperz}
\begin{tabular}{llcc}
\toprule
src & Group & n papers & perc all papers \\
\midrule
top contributors & top\ 30 first\ author\ country & 54,815 & 78.6\% \\
top contributors & top\ 30 journal\ name & 7,883 & 11.3\% \\
top contributors & top\ 30 first\ author\ inst & 2,474 & 3.5\% \\
top contributors & top\ 30 first\ author\ names & 416 & 0.6\% \\
publication type & Basic/Other Research & 43,898 & 62.9\% \\
publication type & Review/Meta-Analysis & 8,725 & 12.5\% \\
publication type & Randomized Control Trial & 8,341 & 12.0\% \\
publication type & Case Report or Comparative Study & 4,530 & 6.5\% \\
publication type & Clinical Trial & 3,121 & 4.5\% \\
publication type & Observational Study & 1,130 & 1.6\% \\
\bottomrule
\end{tabular}
\end{table}


\zsubsection{Thematic structure, coherence and temporal evolution.}

In this section we map the thematic structure of the field by identifying the dominant and recurrent topical themes, aggregating them into higher-order thematic categories, and assessing inter-theme relationships using network-based measures.
This analysis provides a coherent view of the topics attracting research attention and offers insights into theme emergence, persistence, and reconfiguration across the identified developmental epochs of the field.

\begin{figure}[ht!]
	\centering
	\includegraphics[width=0.85\textwidth]{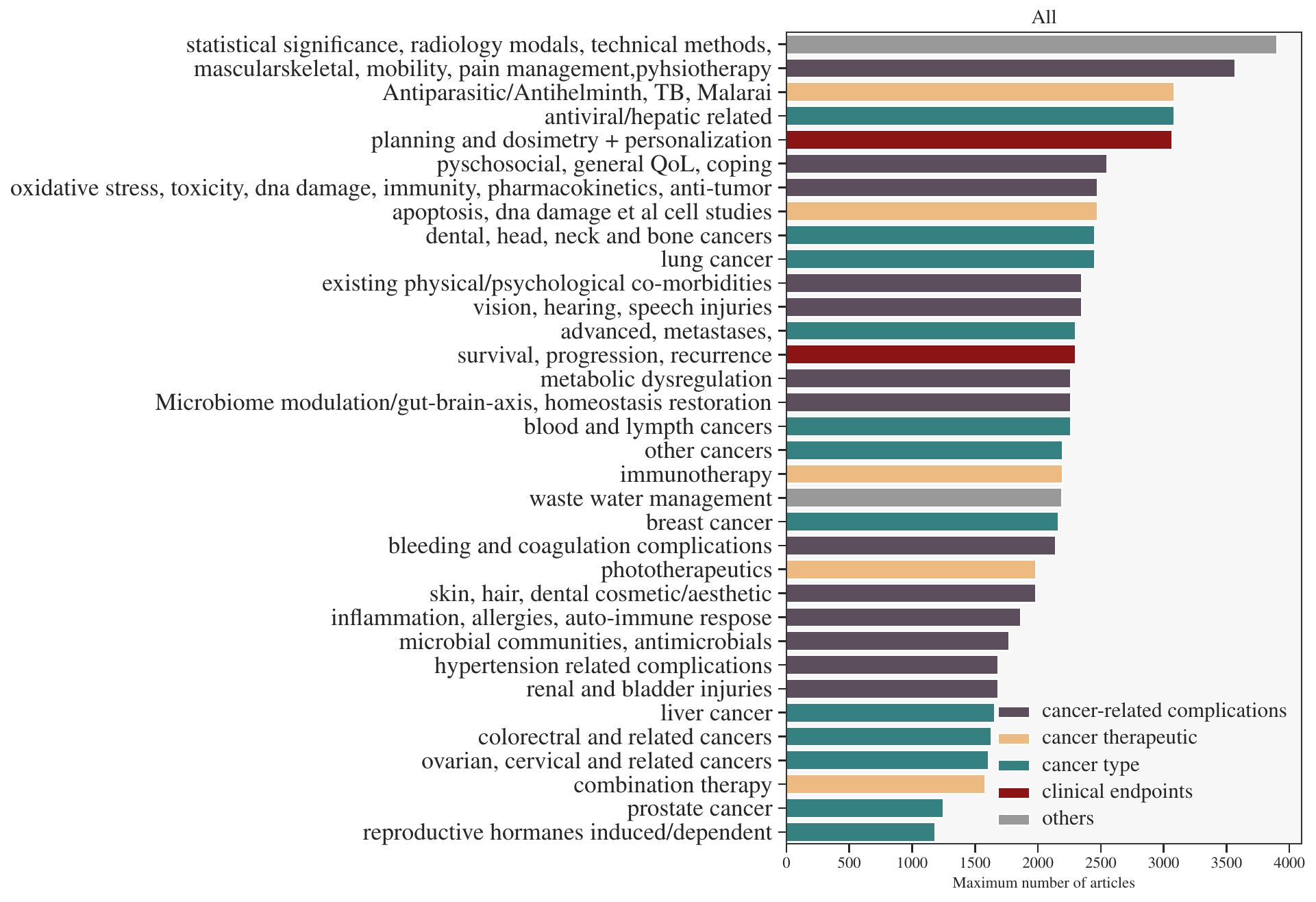}
	\caption{List of identified themes}
	\label{fig:themez-list}
\end{figure}

\zsubsubsection{Prominent themes.}
\Cref{fig:themez-list} illustrates the thematic areas identified through our topic modeling and aggregation pipeline. These themes are determined with a high degree of agreement across the topic modeling runs, with a Jaccard thresholding coefficient of ~\JaccardThreshold. They represent a stable and reproducible set of recurrent thematic emphases in the literature. 
In the figure, themes are ordered on the y-axis by the maximum number of documents linked to each topic across the modeling runs since individual documents may span multiple topics.
For interpretive clarity, we post-hoc aggregate the themes (without altering the underlying topic assignments) into five higher-order categories:
1.~\TScopeCancerTypes – themes specifying cancer types or anatomical tumor sites;
2.~\TScopeComplications – treatment-related adverse events, radiotherapy-associated toxicities, or disease-related complications;
3.~\TScopeCTherapeutics – anticancer therapeutics and biological mechanisms, including immunomodulation and drug repurposing; 
4.~\TScopeOutcomes – clinical outcomes, patient-reported outcomes, treatment planning, and individualized treatment strategies
5.~\TScopeOthers – additional topics such as methodological considerations.

The thematic landscape spans a broad spectrum of tumor entities and anatomical tumor sites, 
including lung, head and neck, dental, bone, breast, liver, colorectal, ovarian, cervical and prostate cancers. Additionally, locally advanced and metastatic disease stages,
and virus-associated or treatment-induced hepatocarcinogenesis
receive particular attention. 
The scope of complications investigated is equally broad, encompassing 
symptom burden, sensory and functional impairment, and health-related quality of life,
such as 
pain and mobility impairment, psychosocial health and overall quality of life, 
vision and speech deficits, gut microbiome disturbances (dysbiosis), metabolic dysregulation, coagulopathies and bleeding (hemostatic abnormalities), dermatologic and hair toxicities, inflammatory and allergic reactions, renal injury, and hypertension. 
Moreover, studies examine TCM applications alongside antibiotic or antimicrobial use and pharmacological investigations related to oxidative stress, toxicity, DNA damage, and anti-tumor activity. 
Therapeutic research covers combining radiotherapy with immunotherapy, 
molecular and cellular mechanisms, including 
apoptosis, DNA damage, toxicity and tumor control, 
as well as repurposing non-oncologic agents from antiparasitic, antihelminthic, antimycobacterial and antimalarial drugs in cancer models. 
Finally, clinical outcomes include key endpoints such as overall survival and progression-free survival, as well as treatment planning and delivery concepts like dosimetry and individualized treatment planning. 
Methodological rigor is reflected in transparent statistical reporting, and integration with radiology imaging modalities suggesting imaging-based biomarkers or radiomic endpoints for evaluating treatment response.

\begin{table}[htbp!]
\centering
\caption{Global network properties of the thematic co-occurrence networks}
\label{tbl:themez-net-globalz}
\begin{tabular}{lcccc}
\toprule
metric name & Low-Alignment & First Wave & Second Wave & All \\
\midrule
n components & 1 & 2 & 4 & 1 \\
network density & 0.187 & 0.305 & 0.177 & 0.301 \\
\bottomrule
\end{tabular}
\end{table}

\begin{figure}[htbp!]  
    \centering  
    \includegraphics[width=0.95\textwidth]{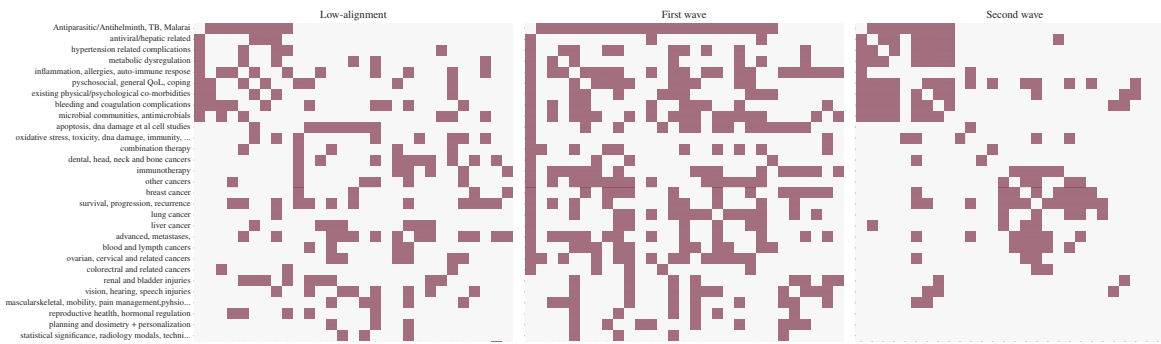}
    \caption{Thematic co-occurrence networks}
    \label{fig:themez-adj-heatmaps}
    \end{figure}

\zsubsubsection{Evolution of thematic emphases}
The conceptual co-occurrence relationships among research theme are shown in the co-occurrence networks in~\Cref{fig:themez-adj-heatmaps}, represented as binary adjacency matrices in which a value of 1 denotes a shared bibliographic reference between themes. 
These networks are stratified by the identified epochs of positive growth, and complemented by associated global-level network analysis statistics in~\Cref{tbl:themez-net-globalz}. 
Thee literature shifts from broad thematic exploration to more focused concentration in the recent epoch. 
During the~\EpochNascent~period, themes are moderately interconnected with a network density of $0.187$ over a single network component. This interconnection of themes intensifies in the~\EpochFirstWave~epoch where the network density rises to $0.305$ (two network components). 
In the~\EpochSecondWave~, however, the network density drops to $0.177$ and the number of connected components increases considerably to $4$. This epoch has visible clustering of themes in the adjacency matrix, which correspond to specialized research domains.  
While this suggests specialization and exploration of new areas, it also reflects an emerging compartmentalization of the field.

\begin{figure}[htbp!]  
    \centering  
    \includegraphics[width=0.95\textwidth]{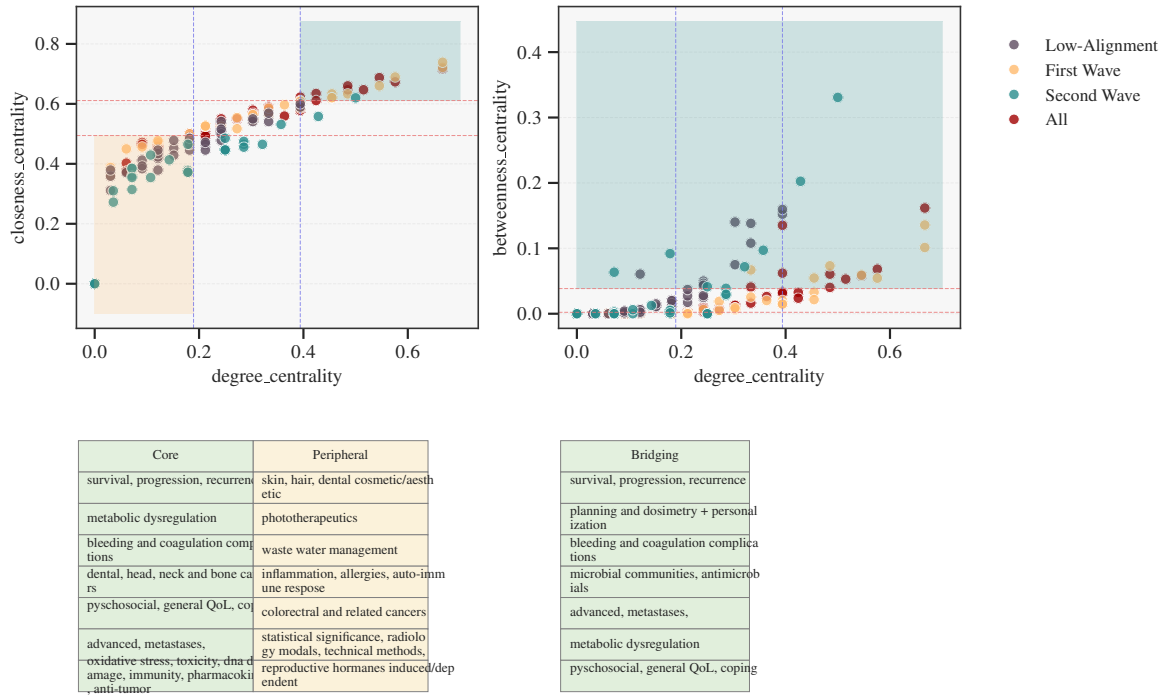}
    \caption{Identification of core, peripheral, and bridging themes}
    \label{fig:themez-net-analysis}
    \end{figure}

\zsubsubsection{Structural concentrations of research themes}
We analyzed the thematic networks to categorize research topics according to their structural roles, identifying three classes: Core, bridging, and peripheral (niche) themes. 
Core themes, marked by high centrality values, reflect the field's dominant and most extensively developed areas in terms of publication volume. Bridging themes exhibit high betweenness centrality and facilitate conceptual linkage across subfields. 
Peripheral themes, defined by low centrality, represent low-connectivity or specialized areas of investigation. These categories are summarized next and their classification is modeled in~\Cref{fig:themez-net-analysis}.
Each subplot in~\Cref{fig:themez-net-analysis}~displays degree centrality against a second network metric and has inter-quartile boundaries (25th and 75th percentiles) delineated by dashed vertical and horizontal lines. 
This network-based partitioning of the themes identifies the central themes occupying the three structural roles.

In the first subplot, which compares degree and closeness centrality, themes in the upper-right quadrant (beyond both 75th percentiles, shaded~\colorCoreThemez) constitute structurally central research themes herein referred to as the foundational or core areas. Conversely, themes in the lower-left quadrant (below both 25th percentiles, shaded~\colorNicheThemez) reflect peripheral or highly specialized domains. 
In a similar fashion, the second subplot identifies the bridging themes using betweenness centrality, while the other two subplots reiterate the structural role of the core and niche themes using eigenvector centrality and clustering coefficient metrics. 

\paragraph{\ThemesCoreCAP.} 
These themes constitute the most extensively connected areas in the network, indicating that a substantial proportion of research activity on TCM as an adjunctive modality to radiotherapy engages with one or more of them.
They include 1. Clinical endpoints such as overall survival and progression-free survival, 2. Management of therapy-related adverse events and supportive care outcomes such as coagulopathies, metabolic dysfunction, and psychosocial wellbeing, 3. Locally advanced and metastatic disease presentation, and cancers of oral (dental), head, neck and bone, and 4. Pharmacological and molecular mechanism investigations. 
Collectively, these foundational areas emphasize patient outcomes, overall wellbeing and mechanistic inquiry. In addition, advanced-stage or systemic disease and bone-related cancers appear to be the frequently studied or early-evaluation contexts for adjunctive TCM use.

\paragraph{\ThemesBrokersCAP.} 
Most core themes reappear in the betweenness centrality analysis, while additional bridging domains include treatment planning and individualized radiotherapy, and management of cancer-associated infections and antimicrobial use. 
The most conceptually central topics in the field also serve as primary integrators of other research domains, anchoring both the volume and flow of knowledge. 
In addition, treatment personalization and cancer-related antimicrobial infections or resistance inherently span diverse data streams, clinical objectives and cancer types, and, therefore, point to cross-cutting operational domains connecting the themes.

\paragraph{\ThemesNicheCAP.} 
These themes address colorectal and hormone receptor-associated cancers; adverse effects involving skin, hair, inflammation and allergies, and reproductive hormones; and methodological topics such as 
photodynamic or photothermal therapeutic approaches, 
imaging-based assessment or radiologic response evaluation, 
and statistical robustness. 
An environmental or pharmaco-environmental research theme on waste-water management  
emerges during the~\EpochFirstWave~but does not persist into the~\EpochSecondWave.


\zsubsection{Rhetorical framing of reported results}

\begin{figure}[htbp!]
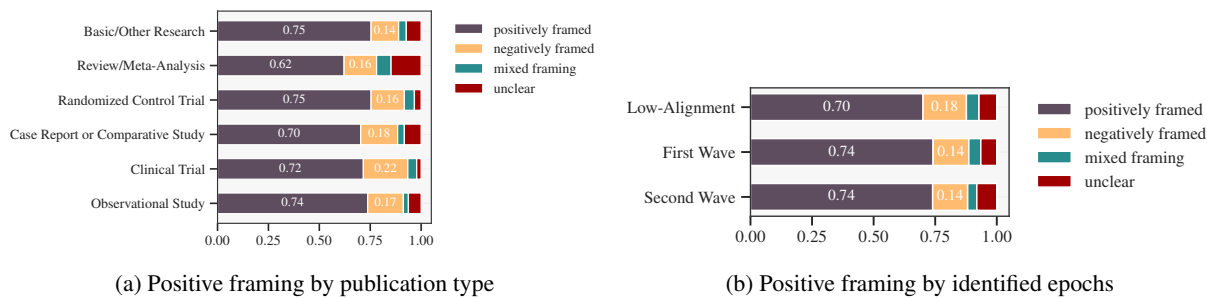
  
    \centering  
    \begin{subfigure}[b]{0.49\linewidth}
        \centering \raisebox{-\height}{\includegraphics[width=\linewidth]{llm-fini\_\_result\_status\_\_study\_types}}
\caption{Positive framing by publication type}
\label{fig:llm-results-by-study-type}
        \end{subfigure}
\begin{subfigure}[b]{0.49\linewidth}
        \centering \raisebox{-\height}{\includegraphics[width=\linewidth]{llm-fini\_\_result\_status\_\_growth\_period\_epoch}}
\caption{Positive framing by identified epochs}
\label{fig:llm-results-by-epoch}
        \end{subfigure}
\hfill
    \caption{Rhetorical framing of results by publication type and identified cycles of evolution}
    \label{fig:llm-results-bars}
    \end{figure}

\begin{figure}[htbp!]  
    \centering  
    \includegraphics[width=0.95\textwidth]{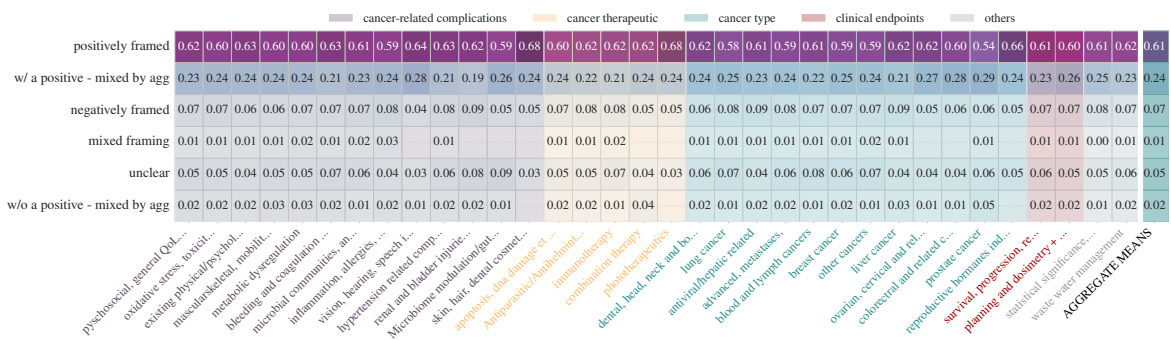}
    \caption{Rhetorical framing of results by identified thematic areas}
    \label{fig:llm-results-heatmap}
    \end{figure}

To capture how the field communicates research progress, we analyzed the framing of results within abstracts, classifying each extracted result statement as 
positively framed, negatively framed, mixed framing or indeterminate framing. 
This classification reflects rhetorical expression of findings, rather than clinical effectiveness or biological efficacy, and provides a lens into the prevalence of positively valenced result reporting. 
Consequently, it maps potential patterns of selective outcome reporting and optimism bias.

The labels \textit{\lblSuccessful}, \textit{\lblUnSuccessful} and \textit{\lblMixed} indicate that a result statement explicitly reports a successful, unsuccessful or mixed result respectively, while the label \textit{\lblUnclear} captures cases where the classifier could not confidently determine the appropriate class. 
This classification is applied to each reported result statement in an abstract, allowing for multiple result statements per document. 
An LLM-based extraction and classification pipeline, filtered through an adjudicator LLM and strict exact-matching validation, is employed, 
ensuring extractive fidelity and semantic faithfulness to source abstracts. 
This process is detailed in the methodology section and associated supplementary material.

To provide further resolution, we stratify the analysis of this label by publication type, distinguishing translational and clinically-oriented studies from basic research investigations and leveraging PubMed's article type classification. We additionally analyze the label by thematic area, 
associating patterns of rhetorical framing to specific domains of inquiry.

\paragraph{Validation of LLM output} 
Validation of the extracted result statements exhibited substantial performance differences between semantic adjudication by the LLM and classical string-based exact matching. 
The LLM judge reported $94\%$ extractive fidelity of extracted result statements and $96.4\%$ label accuracy. Conversely, the exact-matching method determined only $52.7\%$ of the classification to be correct. 
Adopting a conservative precision-oriented position, we  restrict downstream analyses to records verified by exact-matching. 
The resulting analytic subset comprises~\NOBSLLMVALIDATED~extracted result statements from ~\NPMIDZLLMVALIDATED~publications (~\NPMIDZLLMVALIDATEDPERCENTALL~of publications in entire dataset).

\paragraph{Distribution of positively framed result statements.}
The distribution of the rhetorical success classification of reported results indicates a predominance of positive reporting for both the identified developmental epochs and  the publication types. 
Across the three growth epochs (\Cref{fig:llm-results-by-epoch}), at least $70\%$ of the result statements are positively framed compared to at most $18\%$ being negatively framed statements of findings. 
As for the study types (\Cref{fig:llm-results-by-study-type}), 
translational-oriented studies exhibit positively framed result rates comparable to basic research, with the highest proportion of positively framed result statements 
observed in randomized controlled trials (RCTs) and~\lblBasicResearch~at 75\% each.  
Conversely, the clinical trial publication type reports the highest rate of negatively framed result statements at $22\%$, which is also higher than the $14\%$ negatively framed reporting rate within basic research.

\paragraph{Distribution of rhetorical framing by thematic domain.}
Similarly, we observe limited thematic heterogeneity in rhetorical positivity. 
~\Cref{fig:llm-results-heatmap}~presents document-level aggregation of rhetorical framing across thematic areas. 
Unlike the prior analysis, where the observation unit is a result statement, thematic analysis is at the document level, necessitating aggregation. 
As a result two aggregation labels are introduced: 
1. Documents with mixed rhetorical framing including at least one positively framed result statement, and 
2. Documents with mixed rhetorical framing without any positively framed finding. 
On average, the various thematic areas are composed of $61\%$ positively framed findings, $9\%$ negatively framed results ($7\%$~\lblUnSuccessful~and $2\%$ mixed without any positively framed result), and $25\%$ mixed framing results ($1\%$ single mixed result and $24\%$ with at least one~\lblSuccessful~finding).

\section{Discussion}
\label{sec:ddiscuss}

Our study examines global scholarly and clinical research output on TCM as an adjunctive modality to radiotherapy within integrative oncology, where TCM is positioned relative to standard-of-care oncological treatments in the pre-treament, concurrent, or post-treatment phases of conventional cancer care.
Using a longitudinal, multi-dimensional analytical framework that combines standard bibliometric indicators, temporal growth patterns, unsupervised topic modeling, co-occurrence-based thematic network analysis, and analysis of rhetorical framing of reported findings, we assess how the research orientation and thematic scope of the literature has developed over the past twenty-five years.

\paragraph{Growth and visibility in mainstream research.}
Analysis of publication output indicates that research on TCM as an adjunctive therapy to radiotherapy is experiencing growth, a pattern that is also identified in earlier bibliometric analyses~\cite{bai_bibliometric_2023,lian_bibliometric_2025,lei_bibliometric_2025}.
The field shows an annual growth rate of~\PercAnnualGrowthPublications~in publication volume, with 64.9\% of total publication output concentrated in the last decade.

Evidence of growing integration into mainstream research discourse is reflected in the diversity of publication venues. Research on TCM-adjunct oncology appears across multi-disciplinary journals alongside specialist oncology and method-focused outlets. 
Such cross-disciplinary dissemination and visibility in high-impact journals is typically interpreted as a marker of mainstream integration and epistemic legitimacy in metascience studies~\cite{bai_bibliometric_2023,lian_bibliometric_2025}. 
Concurrently, the field is characterized by multi-institutional and international collaboration, which additionally reflects the interdisciplinary nature of the field.

\paragraph{Stability and coherence of the thematic structure.} 
Using an unsupervised topic modeling framework designed to prioritize robustness, we identify a stable and internally coherent thematic structure in the literature on TCM as an adjunctive modality to radiotherapy. A high cross-run agreement across ~\nPermutationsThemesActual~independent topic modeling instantiations is set with a Jaccard similarity-based threshold of 98.5\%. This enables the identification of themes robust to model specification choices and stochastic initialization. 

Five high-level thematic domains emerge: cancer-type-specific foci; supportive and symptom management in oncological care; therapeutics and mechanistic investigations; clinical endpoints, outcomes, and treatment planning; and methodological and analytical considerations. 
In addition, while the increased thematic fragmentation in the most recent temporal epoch (the~\EpochSecondWave) points to topic specialization and thematic concentration, it also signals potential development of epistemic or thematic silos.

\paragraph{Cyclical growth of the research field.}
Synchronous year-to-year variation across multiple bibliometric and research activity indicators (publication output, collaboration breadth, and reported funding acknowledgments) reveals cyclical temporal dynamics in the evolution of the field.
Unlike in prior works where the phases of development are predefined~\cite{zhang_evolution_2023}, in this study, the epochs of expansion emerge inductively from the empirical data.
The field advances through multi-year phases of coordinated expansion across indicators punctuated by shorter periods of deceleration.
On the basis of these aligned dynamics, we identify three broad epochs: an early \EpochNascent~period, a \EpochFirstWave~of coordinated expansion, and a more recent \EpochSecondWave~ of increased thematic focus and research specialization. 
The alignment of multiple indicators suggests that these cycles reflect a shared response to structural and institutional features of the research ecosystem.

Changes in thematic co-occurrence networks across these empirically derived temporal phases of field development provide additional empirical grounding. 
Early research activity is organized around a lower-density and weakly clustered thematic structure, which gives way to a highly interconnected and cross-cutting network during the first expansionary phase. This subsequently transitions to a more fragmented structure in the second wave. 
For interpretive clarity, we draw a heuristic analogy to the iterative \emph{define-ideate-test} framework from design thinking, which is commonly used to describe iterative problem-framing and solution-generation~\cite{dorst_core_2011}. 
We liken~\EpochNascent~epoch to exploratory problem framing under high epistemic uncertainty, \EpochFirstWave~to expansive ideation, and \EpochSecondWave~to hypothesis testing and incremental refinement through thematic specialization. 
Although metascience-based analysis captures aggregate behavior arising from many independent actors and is unlike the planned and facilitated process of design thinking, this macro-level structural correspondence between temporal growth phases and thematic structure is nonetheless striking. We reiterate that this analogy is purely illustrative and identifying causal mechanisms underlying this alignment calls for further investigation.

\paragraph{TCM integration is patient-centered and systems-oriented.}
The thematic structure of the literature indicates that TCM is integrated across multiple levels of oncological care delivery. 
Reflecting both the holistic and syndrome-based (zheng) foundations of TCM, as well as the aims of integrative oncology, particularly symptom management and supportive care, the literature positions TCM within supportive care, symptom management, and quality-of-life-oriented care contexts, and across multiple cancer sites and disease contexts. This is additionally evidenced in the resulting set of core and bridging themes, where supportive care, patient-reported and patient-centered clinical endpoints, and cross-cutting clinical and operational domains persist alongside pharmacokinetic (PK), pharmacodynamics (PD), and mechanistic investigations. 
Together, these patterns suggest that research on TCM as an adjunctive to radiotherapy emphasizes patient well-being, health-related quality of life, and clinical endpoints alongside methodological rigor in study design and analysis. 

Integrative oncology is defined as a patient-centered, evidence-informed model of care that integrates therapies and natural products drawn from diverse medical traditions, including TCM practices, alongside conventional treatments~\cite{cramer_integrative_2013,witt_comprehensive_2017,mao_integrative_2022, gowin_integrative_2024}.

\paragraph{Rhetorical framing of results.}
We additionally assess the rhetorical framing of reported findings by classifying abstract-level result statements as positive, negative, mixed, or indeterminate outcome framing.
This provides a complementary perspective on how findings in research on TCM as an adjunctive in oncology are communicated.
We observe a predominance of positively framed reporting that does not systematically vary across identified cyclical growth phases,
thematic areas, or translation-oriented versus basic or preclinical research publication types. 
Since the analysis captures patterns of reported positivity rather than clinical or mechanistic efficacy, our observed predominance of positively framed results should be interpreted as a systematic characteristic of abstract-level reporting practices, rather than as evidence of therapeutic effectiveness. 
Moreover, this pattern of positive reporting is consistent with documented tendencies toward selective outcome reporting and optimism bias in scientific abstracts~\cite{kavvoura_selection_2007,fanelli_negative_2012,vinkers_use_2015,duyx_strong_2019}. 
The lack of theme-specific heterogeneity in reporting polarity
further suggests that positive framing is a domain-independent property of reporting within TCM adjunctive oncology research.


\subsection*{Limitations}

The "Define-Ideate-Test" cycle described in design thinking literature is employed in this study as a heuristic interpretive analogy. While, in this context, reference to it is informed by patterns observed in the quantitative thematic network analysis, we present it only as an analogy to illustratively describe macro-level patterns in the temporal evolution of the field. 

Additionally, despite the large size of our dataset, the analysis remains bounded by the coverage characteristics of PubMed/MEDLINE, and Chinese-language publications indexed outside of MEDLINE are not represented. 

While we implement author name disambiguation, we caution the challenge of name homonym in East Asian authorship. 
In addition, our disambiguation approach links author identities to institutional affiliations at the time of publication, which may lead to fragmentation of author publication records across career stage or affiliation changes. 

Moreover, the final step of our thematic modeling process, after determination of a stable thematic structure across multiple modeling instantiations, is a curated step that might be influenced by author perspectives.

\section{Materials and methods}
\label{sec:dpmethod}

This study applies a structured computational bibliometric and text-analytic framework to assess the annual publication output of the field, the latent thematic structures, and the rhetorical framing of reported findings. 
The methodology comprises five analytical modules: 1). Data acquisition, curation and preprocessing, 2). Iterative thematic extraction and stability refinement, 3). Descriptive and longitudinal bibliometric profiling, 4). Network-based modeling of thematic co-occurrence and linkage structures, and 5). Large language model-based extraction and classification of abstract-level rhetoric of reported publication findings. 
Each step is described in detail with emphasis on computational reproducibility and  procedural validation, and an end-to-end analytical workflow is presented in~\Cref{fig:thematic-process}.

\input{methdology-workflow}


\zsubsection{Study design and data source}
We perform a systematic bibliographic search of PubMed for publications examining TCM as an adjunctive or supportive modality to radiotherapy, and published from 1 January 2000 onward.
The search is constructed to capture the intersection of TCM, radiotherapy and peri-treatment supportive care, and includes all English-language publication types indexed in PubMed.
Records are retrieved using NCBI EDirect utilities and supplemented by citation-based expansion using references linked on PubMed Central. 
The complete strategy is reported in ~\Cref{sec:appendix-search}. 

Data extraction, curation and normalization entails removing duplicates, harmonizing metadata fields, resolving author-name ambiguities and standardizing author affiliation country fields. Additionally, we integrate author-provided keywords and MeSH controlled vocabulary terms, and derive bibliometric indicators including authorship and funding acknowledgment counts. 
~\Cref{fig:thematic-process} summarizes the data search, retrieval and post-processing workflow.


\subsection{Classification and validation of reporting valence}
We implement a structured computational pipeline to classify the directionality of abstract-level reported findings. This enables us to determine structural patterns in preferential reporting of positively framed findings regarding TCM as a supportive modality in radiotherapy. 
The approach involves extracting abstract-level result statements using~\LLMClassifierFULL~LLM, and subsequently assigning reporting-valence categories to each extracted statement. 
There are four categories of reporting valence:~\lblSuccessful,~\lblUnSuccessful,~\lblMixed, and~\lblUnclear. 
The categories~\lblSuccessful,~\lblUnSuccessful, and~\lblMixed, correspond to direct statements of successful, unsuccessful or mixed findings,
while the category~\lblUnclear~reflects cases with low classification confidence. 
Individual abstracts can contain multiple result statements and each is treated as an independent unit of analysis.

From the full corpus generated during search and post-processing, we draw a random sample of $5,000$ records. The sample size is constrained by computational resource requirements for LLM inference. 
Validation of extracted abstract-level result statements and their reporting-valence categorization employs two complementary validation strategies. 
First,~\LLMClassifier~is used for automated adjudication to assess the extraction fidelity of each statement, and the consistency between extracted statements and assigned reporting-valence labels. 
Second, we apply rule-based pattern-matching using conventional natural language processing techniques to verify the correctness of extracted statements. 
The final classification accuracy estimate is taken as the conservative accuracy estimate defined as the minimum of the two validation scores.


\zsubsection{Modeling of thematic areas} 
To identify the thematic structure within the corpus, we use embedding-based semantic topic modeling, which leverages contextual semantic representations and allows modeling at the document-representation level, without reliance on bag-of-words or n-gram dimensionality reduction. As a clustering-driven approach, stochastic variability across model initializations and hyperparameter settings can occur. Therefore, we construct a pipeline designed to reduce stochastic instability and parameter sensitivity
(\Cref{fig:thematic-process}). 
Each topic modeling run is characterized by a document form, a text embedding model and clustering hyperparameters. 
The input document is constructed as a concatenation of the publication title, abstract, author keywords, and MeSH terms. The employed pretrained transformer-based language embedding models are~\BERTFull~and~\QWEN, and the principal component dimensionalities prior to clustering are $ \in {13, 20, 31, 65, 130, 200, 310}$, with the full embedding size being $768$.  

For each configuration, we generate up to $20$ thematic clusters, where a thematic cluster is composed of the 10 top-ranked terms by cluster relevance.  
To consolidate similar themes, we compute pairwise Jaccard similarity coefficients ($\frac{|A n B|}{|A u B|}$) on the discovered terms and apply a~\JaccardThreshold~similarity threshold for theme aggregation. 
The final thematic sets are then manually curated and assigned descriptive labels that reflected their underlying term distributions. 

Furthermore, thematic co-occurrence matrices are derived from document level co-assignment and graph-theoretic network metrics computed to characterize node centrality and network cohesion measures. 
Convergence across multiple centrality measures informs the identification of~\ThemesCore,~\ThemesBrokers, and~\ThemesNiche thematic areas.

\zsubsection{Indicators and analysis methods}
We examine several bibliometric indicators of research activity, including publication volume, authorship counts, institutional affiliations, country affiliations, and funding acknowledgments. 
These indicators are primarily aggregated by publication year. In addition, annual publication counts are cross-tabulated by disambiguated first author identity, journal, and first-author affiliation country and institution.
For longitudinal normalization and comparability, we apply standard z-score transformation to annual publication counts, $\frac{x - \hat\mu(x)}{\hat\sigma(x)}$, where $x$ denotes annual publication count. 
In addition, we compute year-to-year growth rates for each bibliometric indicator as per $\frac{x_{t+1}^{i} - x_{t}^{i}}{x_{t}^{i}}$ , where $x^{i}$ is an indicator and $t$ is time indexed by publication year.

\paragraph{Implementation.}
All computational procedures are implemented in Python using standard scientific computing libraries alongside specialized libraries for natural language processing, network analysis, and machine learning. 
BERTopic~\cite{grootendorst2022bertopic} was used for embedding-based semantic topic modeling, while Bibliometrix~\cite{rbibex} enabled preprocessing of keywords and trigrams.

\section{Conclusion}
\label{sec:dconc} 

This systematic analysis of 25 years of research reveals how TCM has evolved as an evidence-based adjunct to radiotherapy.
The field demonstrates robust growth, with expanding global collaboration and integration into mainstream journals.  
Additionally, it has maintained a focus on patient-centered outcomes (symptom management and treatment tolerance as well as health quality of life). 
The emergent cyclical growth pattern reflects the field's maturation, yet reveals a likely challenge with defragmentation into epistemic silos that may impede the cross-disciplinary aspect of TCM integration into oncology. 
Moreover, positive framing of result is homogeneous across publication types, thematic areas and time, signaling systematic reporting bias across the field.

{
	\bibliographystyle{unsrtnat} 
	\bibliography{\dpaperbib}
}

\newpage 
\onecolumn
\appendix
\renewcommand{\thefigure}{A\arabic{figure}}  
\renewcommand{\thetable}{A\arabic{table}}    
\setcounter{figure}{0}  
\setcounter{table}{0}   

\section{Methods}
\label{sec:appendix-methods}

\zsubsection{Search Method}
\label{sec:appendix-search}

\textbf{Search term}

\begin{lstlisting}[style=pythonstyle, caption={Search Terms}]
	
	(("TCM"[All Fields] AND "Radiotherapy"[All Fields] AND "Pre-treatment"[All Fields]) OR "During treatment"[All Fields] OR "Post-treatment"[All Fields]) AND "2000/01/01:2025/12/31"[Date - Publication]
	
\end{lstlisting}

\textbf{Search results}
Actual Retrieved: Search results n=130,448 w/ duplicates, n=88,363 unique <-- 130k b/c also followed PMC.cites links and 
retrieved those as well. By tracing back through the references to find earlier studies a publication draws upon, we ensure foundational work isn't overlooked. By findings newer papers the cite this publication, we track how the field has built on their results and uncover emerging debates, replication studies and recent methological advances.

Retrieval date for PMIDz list: \PMIDzFetchDate

\zsubsection{Workflow for extracting, classifying and validating reporting valence of result statements}
\label{sec:appendix-llm}

\paragraph{Extract and Classify}

\textit{LLM choice:}~\LLMClassifierFULL

\textit{Prompt Design: }

\begin{lstlisting}[style=promptstyle, caption={Extract and classify reporting valence}]
# SYSTEM PROMPT
	Task instructions and goal
	You're a scientific research assistant in the field of biomedical engineering, and  with excellent entity and relationships recognition skills. 
	Your task is to extract information from a provided text as per the user instructions and `output format` directives. 
	- **Only use facts provided by the user in the `input`/<input> text**  
	
	
	Other guidelines to observe
	- Be concise and succinct
	- if you don't know do not create/imagine facts
	- use the string `**N/A**` for empty non-numeric values  and the number `-99` for empty numeric values if you must report a value and yet you don't know.
	- Striclty adhere to the output format requested. Only extract and return as per the output format or JSON schema; no other detail!! 
	- If JSON output is requested, return only valid json output AND as per the indicated format/schema. **Strictly return valid JSON output and as per indicated output JSON schema**, AND enclose the JSON output in a code block e.g "```json <only requested output as per output JSON schema goes here>```".
	- Where provided, make use of the tools as much as possible; do not ignore the tools or rely on generic information if a relevant tool is availble.
	



# DOCUMENT PROCESSING PROMPT
	Retrieve all results/findings in this text as per the properties in the requested JSON output format. Extract only results/findings that describe the therapeutic effects or clinical outcomes of a treatment or intervention.
		
	<input> \{..... document to process and associated output schema go here .....\} 
	</input>

\end{lstlisting}

\begin{lstlisting}[style=pythonstyle, caption={Extractor schema and classification labels}] 
	class ResultStatus(Enum):
		SUCCESSFUL = "successful"
		NOT\_SUCCESSFUL = "unsuccessful"
		MIXED\_RESULTS = "mixed\_results"
		I\_CANNOT\_TELL = "don't know/unclear"
	
	class AResult(BaseModel):
		result\_statement:Optional[str] = Field(None, description="The reported conclusion result/finding statement as-is.")
		result\_summary:Optional[str] = Field(None, description="A concise and succinct summary of the result in under 30 words.") 
		result\_status:Optional[ResultStatus] = Field(None, description="Is this result a success, failure, mixed results, or cannot tell or don't know.") 
		result\_status\_rationale:Optional[str] = Field(None, description="Brief description (under 30 words) of the reasoning behind the indicated result\_status.") 
\end{lstlisting}

\zsubsection{Aggregation of themes}

\begin{figure}[htbp!]  
    \centering  
    \includegraphics[width=0.95\textwidth]{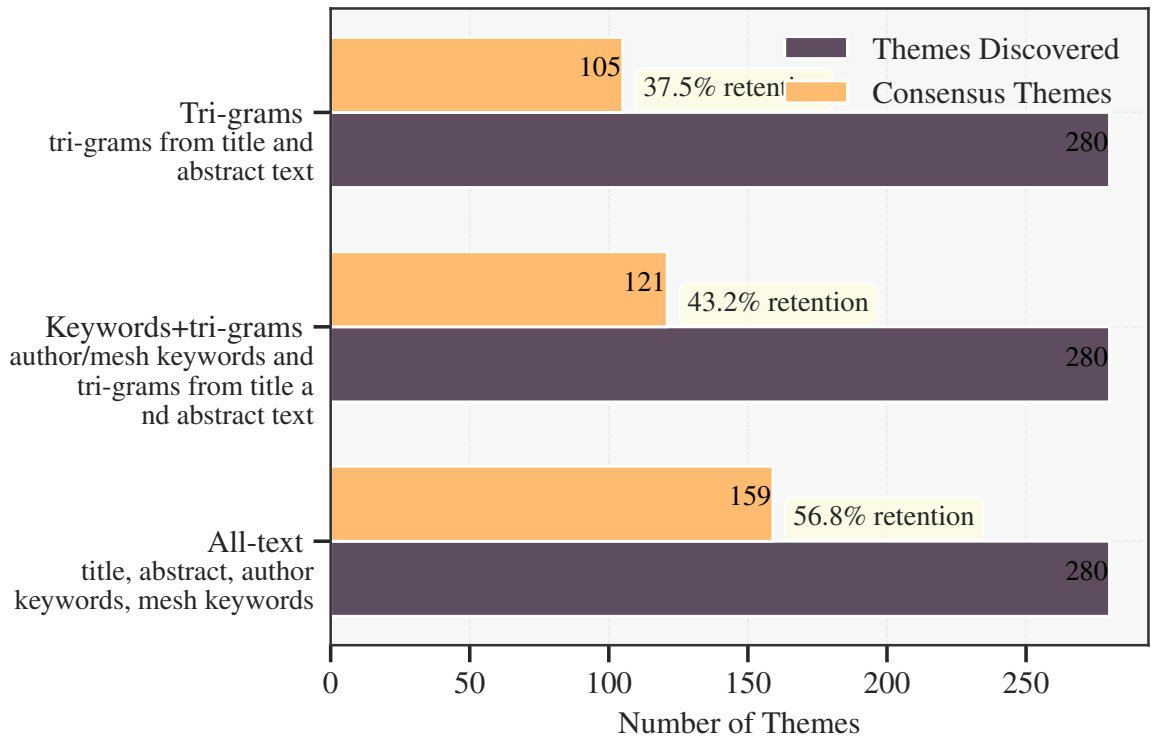}
    \caption{Aggregation of themes from multiple documents and modeling streams}
    \label{fig:themez-consensus}
    \end{figure}

\newpage 
\section{Results}
\label{sec:appendix-results}

\zsubsection{Annual output of contributors}

\begin{figure}[htbp!]
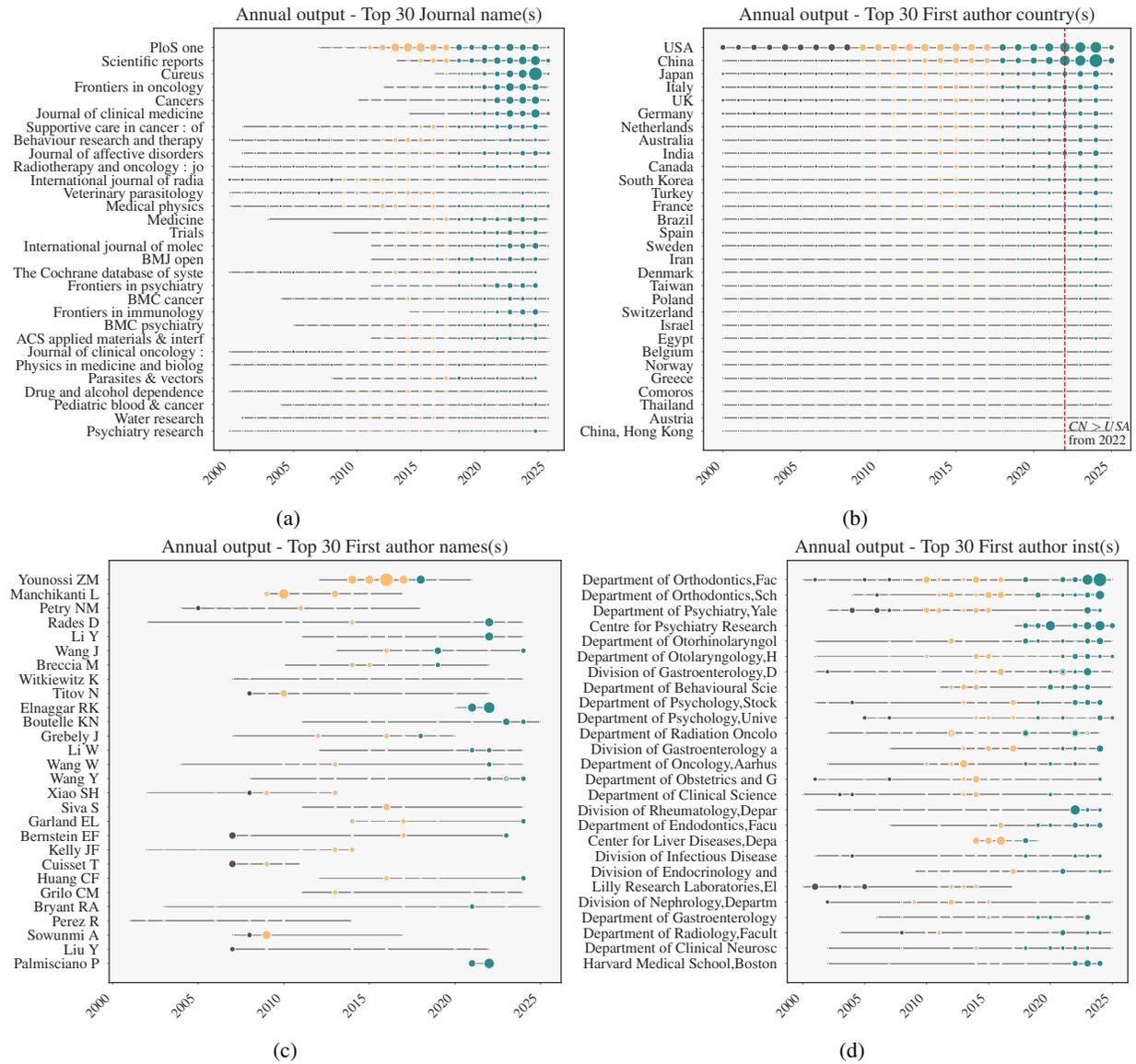
  
    \centering  
    \begin{subfigure}[b]{0.49\linewidth}
        \centering \raisebox{-\height}{\includegraphics[width=\linewidth]{output\_indicators\_\_journal\_name-n\_paperszscore}}
\caption{}
\label{fig:sup-player-journals}
        \end{subfigure}
\begin{subfigure}[b]{0.49\linewidth}
        \centering \raisebox{-\height}{\includegraphics[width=\linewidth]{output\_indicators\_\_first\_author\_country-n\_paperszscore}}
\caption{}
\label{fig:sup-player-countries}
        \end{subfigure}
\hfill\begin{subfigure}[b]{0.49\linewidth}
        \centering \raisebox{-\height}{\includegraphics[width=\linewidth]{output\_indicators\_\_first\_author\_names-n\_paperszscore}}
\caption{}
\label{fig:sup-player-authors}
        \end{subfigure}
\begin{subfigure}[b]{0.49\linewidth}
        \centering \raisebox{-\height}{\includegraphics[width=\linewidth]{output\_indicators\_\_first\_author\_inst-n\_paperszscore}}
\caption{}
\label{fig:sup-player-insts}
        \end{subfigure}
\hfill
    \caption{Annual publication output of various stakeholders}
    \label{fig:output-players-full}
    \end{figure}

\newpage

\zsubsection{Thematic areas}

\begin{figure}[htbp!]  
    \centering  
    \begin{subfigure}[b]{0.99\linewidth}
        \centering \raisebox{-\height}{\includegraphics[width=\linewidth]{69k\_\_nx\_\_degree-centrality-vs-others\_\_75th\_\_grp1}}
\caption{Commonly occuring themes Vs rare/niche}
\label{fig:net-part-1}
        \end{subfigure}
\hfill\begin{subfigure}[b]{0.99\linewidth}
        \centering \raisebox{-\height}{\includegraphics[width=\linewidth]{69k\_\_nx\_\_degree-centrality-vs-others\_\_75th\_\_grp2}}
\caption{Knowledge brokers Vs cliquey}
\label{fig:net-part-2}
        \end{subfigure}
\hfill
    \caption{Identification of core, peripheral, and bridging themes}
    \label{fig:themez-net-analysis-all}
    \end{figure}

\newpage

\begin{table}[htbp!]
\centering
\caption{Roles from analysis of the thematic co-occurrence networks}
\label{tbl:themez-scatter-rolez}
\begin{tabular}{p{1.7in}p{4.3in}}
\toprule
role name & All \\
\midrule
Core/foundational & ['Antiparasitic/Antihelminth, TB, Malarai', 'antiviral/hepatic related', 'bleeding and coagulation complications', 'breast cancer', 'metabolic dysregulation', 'microbial communities, antimicrobials', 'pyschosocial, general QoL, coping', 'survival, progression, recurrence'] \\
Focal & ['Antiparasitic/Antihelminth, TB, Malarai', 'Microbiome modulation/gut-brain-axis, homeostasis restoration', 'antiviral/hepatic related', 'apoptosis, dna damage et al cell studies', 'breast cancer', 'combination therapy', 'existing physical/psychological co-morbidities', 'immunotherapy', 'liver cancer', 'lung cancer', 'other cancers', 'ovarian, cervical and related cancers', 'planning and dosimetry + personalization', 'prostate cancer', 'renal and bladder injuries', 'vision, hearing, speech injuries'] \\
Peripheral/niche & ['Microbiome modulation/gut-brain-axis, homeostasis restoration', 'apoptosis, dna damage et al cell studies', 'blood and lympth cancers', 'colorectral and related cancers', 'combination therapy', 'dental, head, neck and bone cancers', 'immunotherapy', 'inflammation, allergies, auto-immune respose', 'liver cancer', 'lung cancer', 'mascularskeletal, mobility, pain management,pyhsiotherapy', 'other cancers', 'ovarian, cervical and related cancers', 'oxidative stress, toxicity, dna damage, immunity, pharmacokinetics, anti-tumor', 'phototherapeutics', 'planning and dosimetry + personalization', 'renal and bladder injuries', 'reproductive heatlth, hormonal regulation', 'reproductive hormanes induced/dependent', 'statistical significance, radiology modals, technical methods,', 'vision, hearing, speech injuries'] \\
Core and core-adjacent & ['Antiparasitic/Antihelminth, TB, Malarai', 'antiviral/hepatic related', 'bleeding and coagulation complications', 'breast cancer', 'existing physical/psychological co-morbidities', 'metabolic dysregulation', 'microbial communities, antimicrobials', 'oxidative stress, toxicity, dna damage, immunity, pharmacokinetics, anti-tumor', 'pyschosocial, general QoL, coping', 'survival, progression, recurrence'] \\
Core-adjacent only & ['Antiparasitic/Antihelminth, TB, Malarai', 'antiviral/hepatic related', 'existing physical/psychological co-morbidities', 'hypertension related complications', 'microbial communities, antimicrobials'] \\
Bridging & ['Antiparasitic/Antihelminth, TB, Malarai', 'advanced, metastases,', 'bleeding and coagulation complications', 'breast cancer', 'dental, head, neck and bone cancers', 'existing physical/psychological co-morbidities', 'mascularskeletal, mobility, pain management,pyhsiotherapy', 'metabolic dysregulation', 'oxidative stress, toxicity, dna damage, immunity, pharmacokinetics, anti-tumor', 'pyschosocial, general QoL, coping', 'survival, progression, recurrence'] \\
Cliquey & ['Microbiome modulation/gut-brain-axis, homeostasis restoration', 'blood and lympth cancers', 'colorectral and related cancers', 'combination therapy', 'hypertension related complications', 'liver cancer', 'mascularskeletal, mobility, pain management,pyhsiotherapy', 'metabolic dysregulation', 'microbial communities, antimicrobials', 'other cancers', 'ovarian, cervical and related cancers', 'renal and bladder injuries', 'skin, hair, dental cosmetic/aesthetic'] \\
\bottomrule
\end{tabular}
\end{table}

\begin{figure}[htbp!]  
    \centering  
    \includegraphics[width=0.95\textwidth]{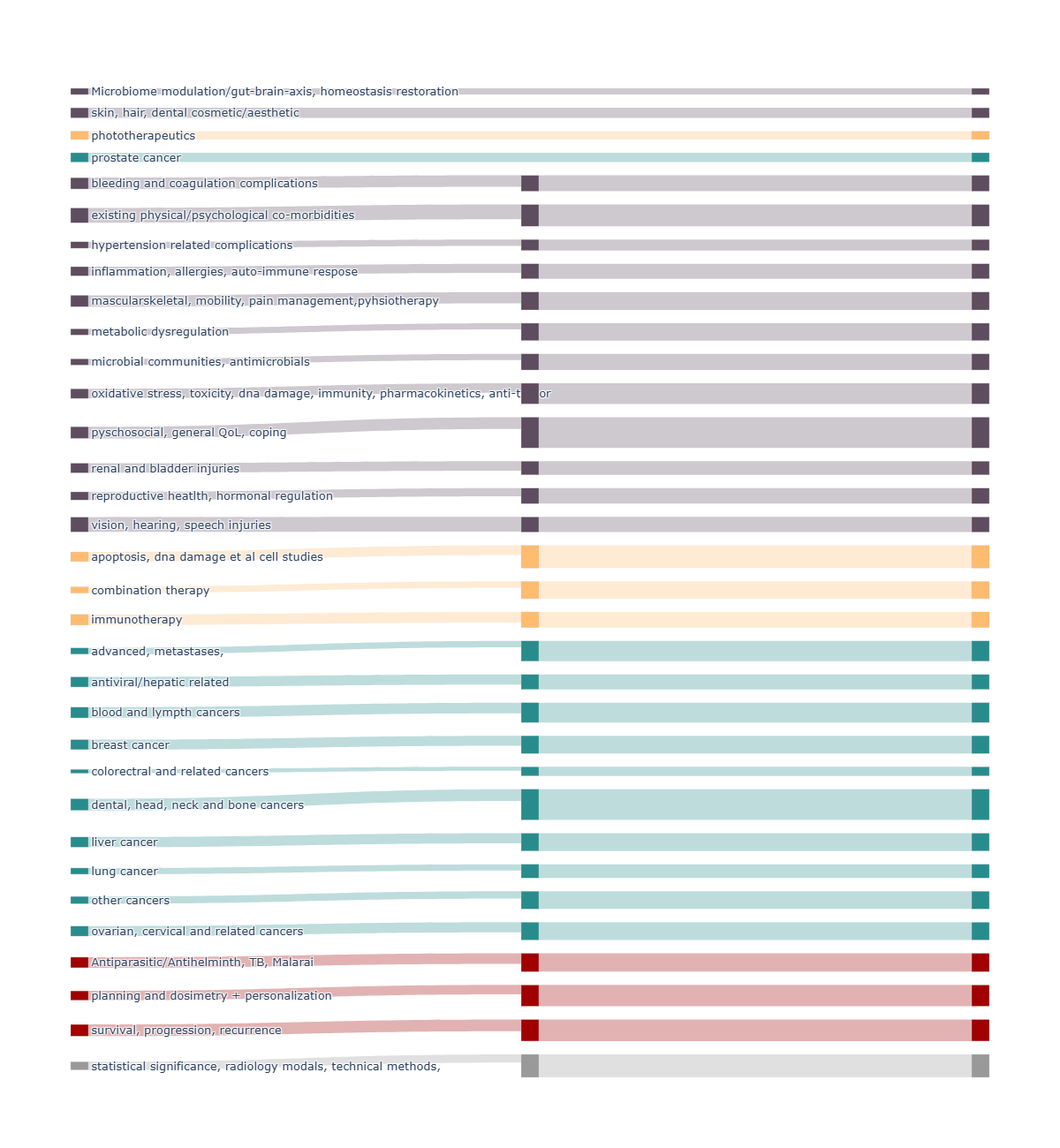}
    \caption{Evolution of themes over time by identified epochs}
    \label{fig:themez-t-sankey}
    \end{figure}

\begin{figure}[htbp!]  
    \centering  
    \begin{subfigure}[b]{0.49\linewidth}
        \centering \raisebox{-\height}{\includegraphics[width=\linewidth]{69k\_\_nx\_\_tepoch\_\_LowAlignment}}
\caption{Low Alignment epoch}
\label{fig:None}
        \end{subfigure}
\begin{subfigure}[b]{0.49\linewidth}
        \centering \raisebox{-\height}{\includegraphics[width=\linewidth]{69k\_\_nx\_\_tepoch\_\_FirstWave}}
\caption{First wave epoch}
\label{fig:None}
        \end{subfigure}
\hfill\begin{subfigure}[b]{0.49\linewidth}
        \centering \raisebox{-\height}{\includegraphics[width=\linewidth]{69k\_\_nx\_\_tepoch\_\_SecondWave}}
\caption{Second wave epoch}
\label{fig:None}
        \end{subfigure}
\begin{subfigure}[b]{0.49\linewidth}
        \centering \raisebox{-\height}{\includegraphics[width=\linewidth]{69k\_\_nx\_\_tepoch\_\_All}}
\caption{All}
\label{fig:None}
        \end{subfigure}
\hfill
    \caption{Network visualization of thematic structures over time by identified epochs}
    \label{fig:themez-netx}
    \end{figure}

\newpage
\zsubsection{Rhetorical Reporting}

\begin{table}[htbp!]
\centering
\caption{LLM validation scores}
\label{tbl:llm-validate-eval-with-laaj}
\begin{tabular}{p{1.in}p{0.6in}p{.3in}p{1.1in}p{.6in}}
\toprule
evaluator & metric name & metric scale & range of values & \% valid \\
\midrule
LLM-as-a-Judge & Factuality & likert4 & Unclear: -99\newline Strongly no: 0\newline No: 1\newline Yes: 2\newline Strongly yes: 3 & 0.964 \\
LLM-as-a-Judge & Faithfulness & likert4 & Unclear: -99\newline Strongly no: 0\newline No: 1\newline Yes: 2\newline Strongly yes: 3 & 0.940 \\
Quantitative ML & Exact match &  & [0-1] & 0.527 \\
\bottomrule
\end{tabular}
\end{table}

\end{document}